\documentstyle[jmlr2e]{article}

\newenvironment{myitemize}
{ \begin{itemize}
    \setlength{\itemsep}{0pt}
    \setlength{\parskip}{0pt}
    \setlength{\parsep}{0pt}     }
{ \end{itemize}                  }

\author{%
    \name Pavel Filonov \email Pavel.Filonov@kaspersky.com \\
    \name Andrey Lavrentyev \email Andrey.Lavrentyev@kaspersky.com \\
    \name Artem Vorontsov \email Artem.Vorontsov@kaspersky.com \\
    \addr{
        Technology Research Department, Future Technologies\\
        Kaspersky Lab \\
        39A/3 Leningradskoe Shosse \\
        Moscow, 125212, Russian Federation 
    }
}

\title{Multivariate Industrial Time Series with Cyber-Attack Simulation: Fault Detection Using an LSTM-based Predictive Data Model}

\editor{}

\begin{document}

\maketitle

\begin{abstract}
We adopted an approach based on an LSTM neural network to monitor and detect faults in industrial multivariate  time series data. To validate the approach we created a Modelica model of part of a real gasoil plant. By introducing hacks into the logic of the Modelica model, we were able to generate both the roots and causes of fault behavior in the plant. Having a self-consistent data set with labeled faults, we used an LSTM architecture with a forecasting error threshold to  obtain precision and recall quality metrics. The dependency of the  quality metric on the threshold level is considered.  An appropriate mechanism such as ``one handle'' was introduced for filtering faults  that are outside of the plant operator field of interest.
\end{abstract}

\begin{keywords}
industrial fault detection, LSTM neural networks, multivariate time series forecast
\end{keywords}

\section{Introduction}

One area that strongly requires a technique for multivariate time series analysis is cyber-security for industrial processes~\citep{Stuxnet}. Conventional cyber-security tools are used to detect malicious activity at the communication level and the binary execution level. Meanwhile, Industry 4.0 and the IoT era means cyber and physical parts are connected in a single Cyber-Physical System (CPS). To protect a CPS one has to use not only conventional means for cyber-security but also perform communication protocol deep package inspection (DPI). A DPI tool needs to monitor and detect faults inside technological processes by analyzing historical and real-time streaming of industrial data.

Numerous approaches to fault detection (FD) in industrial and other types of multivariate time series have been proposed: classic methods like PCA, DPCA, FDA, DFDA, CVA, PLS ~\citep{0957-0233-12-10-706}, SVM and segmentation ~\citep{Lin2007,s150202774}, change point detection ~\citep{MattesonJ13}, LSTM ~\citep{Malhotra:2015, Malhotra:2016, Yadav:2016} ~.

In this paper for the purpose of monitoring and detecting faults inside a multivariate industrial time series that contains both sensors and controls signals, we evolve LSTM-based approach~\citep{Malhotra:2015,Malhotra:2016}.

To validate our approach we needed real object data sets for normal as well as anomalous behavior. Experiments on data sets from several real industrial objects are usually faced with the same problem - the absence of anomalous behavior, or very few examples.  To provide realistic data with anomalies we created a mathematical model of part of a real gasoil plant. Having a model, we were able to modify some of the process logic and generate faults. With a self-consistent mathematical model and knowing the causality relations of model variables, we trained and tested an LSTM neural network and deeply investigated the obtained results and adopted LSTM architecture parameters.

The rest of the paper is organised as follows: Section~\ref{sec:data_desc} describes a data set generated by an industrial process model using different types of attacks. In section~\ref{seq:lstm_fd} we describe an LSTM-based fault detection scheme and consider the results of the experiment. Section~\ref{seq:conc} offers concluding remarks.

\section{Data Set Description}
\label{sec:data_desc}

We created a Modelica model for a gasoil plant heating loop. 

\begin{figure}[h]
    \centering
    \includegraphics[width=\linewidth]{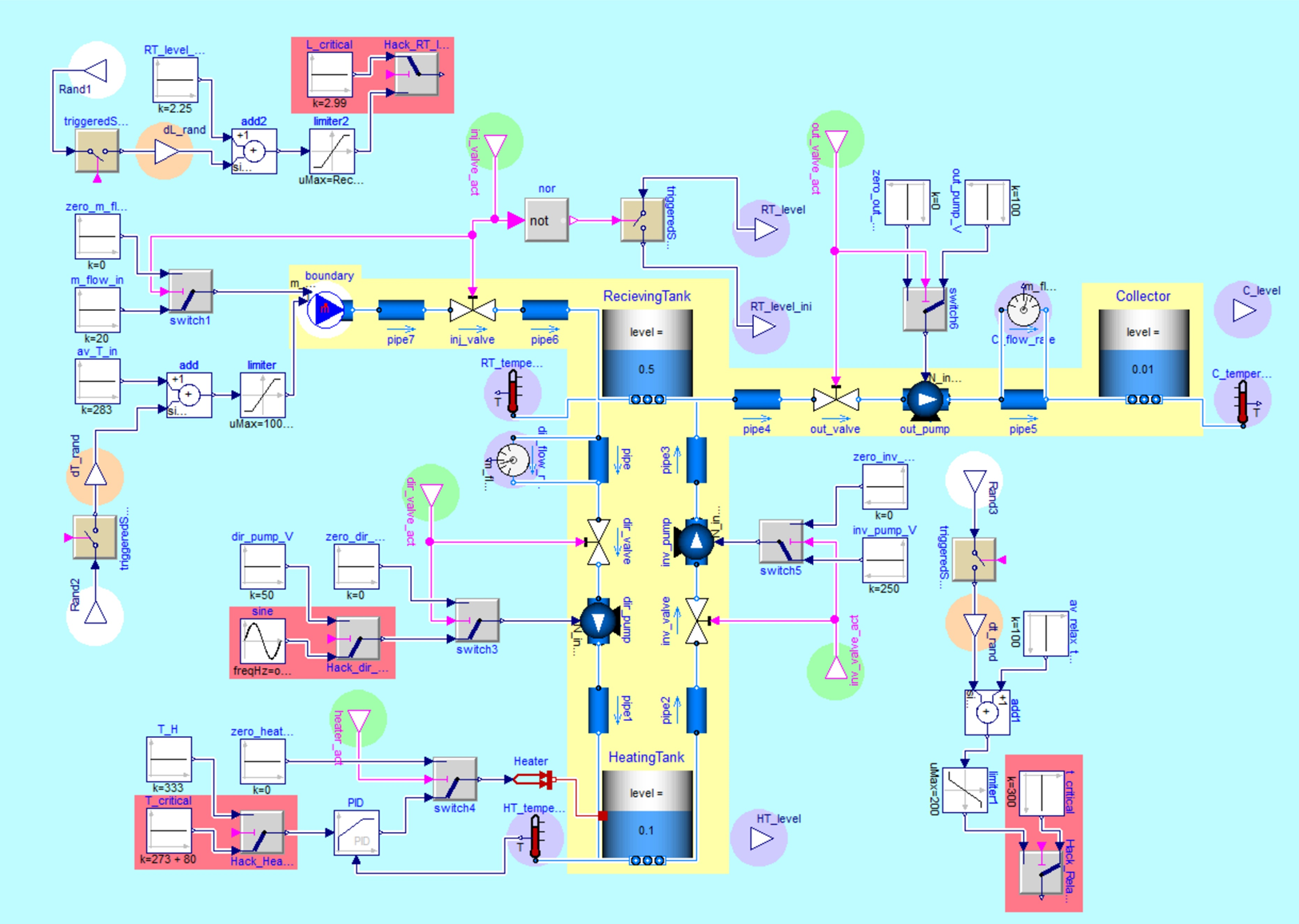}
    \includegraphics[width=\linewidth]{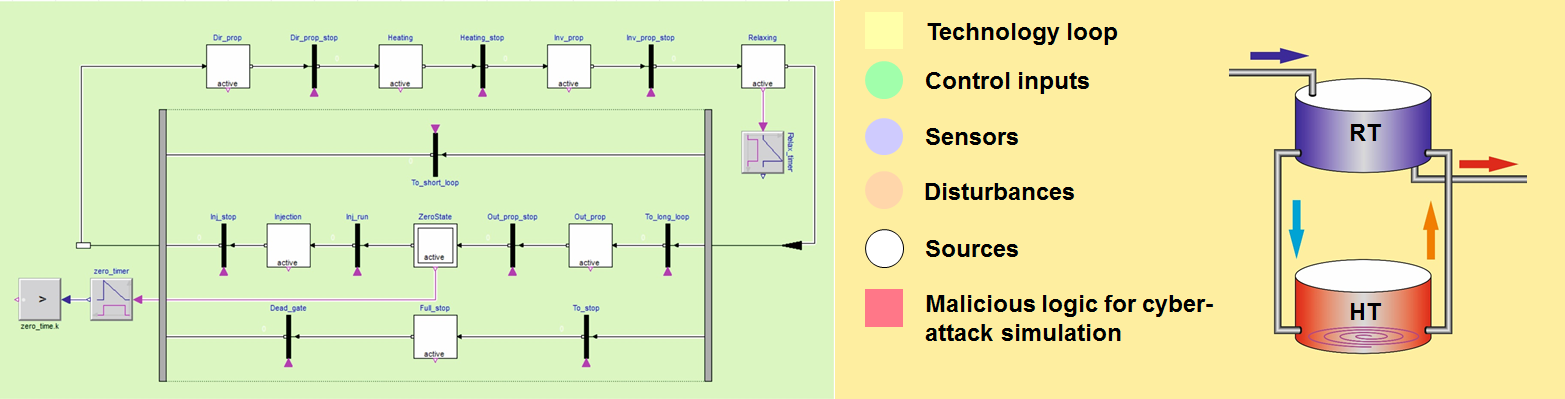}
    \caption{Gasoil Heating Loop Modelica Model}
    \label{fig:modelica}
\end{figure}

The gasoil heating loop~(GHL) model comprises three reservoirs: receiving tank (RT), heating tank (HT) and collector tank (CT). The technological task it to heat gasoil in RT up to 60 degrees Celsius, thus reaching a gasoil viscosity that is enough to transfer it to CT. Heating in the model is performed in portions. A portion of gasoil is heated up to 60 $C$ in HT and then pumped back into RT and ralaxing there for some time. This process is repeated till reaching 60 $C$ in RT. RT is then emptied into CT. After that, RT is refilled from some inexhaustible source.

For simplicity, we used water as the fluid instead of gasoil. We used Dymola to simulate the model.

\begin{figure}[h]
    \centering
    \includegraphics[width=0.7\linewidth]{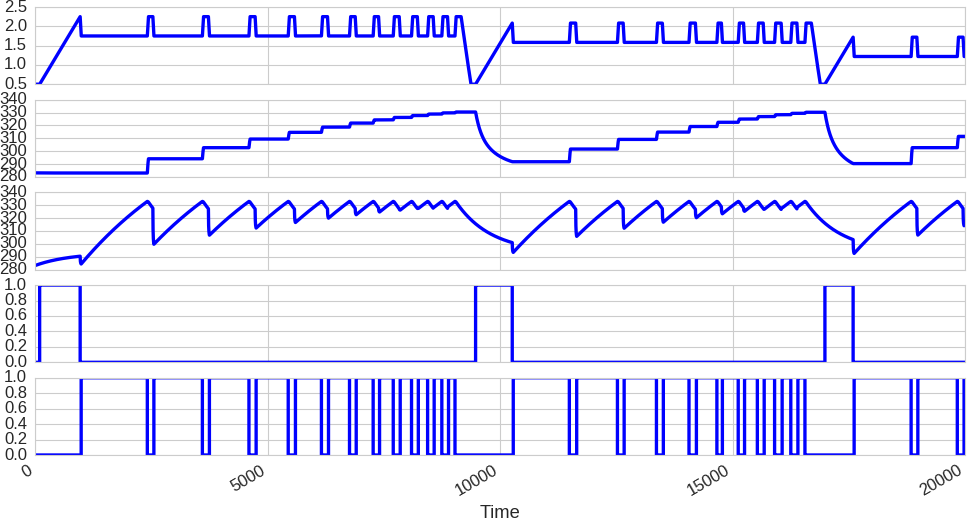}
    \caption{Most important variables (descending): RT{\_}level, RT{\_}temperature.T, HT{\_}temperature.T, inj{\_}value{\_}act,  heater{\_}act. }
    \label{fig:data}
\end{figure}

Using the GHL model we generated a multivariate time series with 270 variables. In this paper we present the results for a multivariate time series with only 19 variables~\citep{GHL22}.  For the complete 270-variables time series we also applied the same technique of fault detection and obtained the same results except that the time for fitting the model was 30\% longer. We selected 19 variables knowing the semantic of data; however, with real object this is not always possible. The most interesting variables of normal behavior are represented in the Figure~\ref{fig:data}. The first three variables are  the sensors of RT level , RT temperature and HT temperature. The last two variables correspond to gasoil source on/off  and heater on/off control signals.

In the GHL model we introduced four types of cyber attack to the normal process logic:
\begin{myitemize} 
    \item unauthorized change of max RT level,
    \item unauthorized change of max HT temperature,
    \item unauthorized change of pump frequency,
    \item unauthorized change of system relaxing time value.
\end{myitemize}

In the current paper we only present the results for fitting and testing the LSTM for anomalies generated by the first type of attack to the max-RT-level set point. By changing the time of attack and the value of the hacked max-RT-level, we generated many anomalous data sets used for fault detection. To train the LSTM we used only a data set with normal behavior.

The generated data has no outlier. When dealing with data from real objects, before learning normal behavior, we perform data preprocessing, thus eliminating outliers and data gaps.

When dealing with cyber attacks at the industrial data level, the main task is to detect anomalous process flows as earlier as possible. 

In the generated data set we know the time when the attacker changed the control logic set-point, the start of the sub-process which is influenced by the attacker’s changes, the time when the sub-process crossed the normal behavior condition and the interval when the attack resulted in an incident. The data-driven model has to ``see'' all of these situations. In a real attack, even if the attacker was able to hide the control logic set-point change event, the data-driven model has to detect a fault at the time-point when the sub-process crosses the normal behavior condition. 

Generated multivariate time series consist of high-dimensional complex nonlinear, non-stationary data with non-Gaussian pointwise distribution. Variables have partially probabilistic nature. Correlations of variables have event-based nature because of control data primacy. As will be shown in the next section accurate fitting of this data using parametric data-driven model requires thousands of model parameters, moreover, complete model learning requires data set containing about million of time points ($\sim 10^6$ sec). Meanwhile temporal evolution of hacker-induced anomalies often are very fast ($\sim 100$ sec) and rapidly grown into an equipment damage. Under these conditions, online process monitoring using traditional change point methods (operating with testing of several statistical hypothesis) are dramatically complicated by the requirement of fast online estimation of thousand model parameters to make decision about starting of anomalous process behavior. Thus without prior information about anomalies and their representation in process trajectories the most appropriate anomaly detection technique operates with fixed model pre-trained on data set under normal operating conditions. Such technique considers anomalies as a deviation of observed process trajectories from trajectories predicted by the model.

\section{LSTM-based Fault Detection}
\label{seq:lstm_fd}

Input data can be described as multivariate time series $X = \{x^{(1)}, x^{(2)}, \dots, x^{(n)}\}$, where $x^{(t)}$ belongs to $m$ dimensional space ${\mathbb R}^m$, $n$ --- number of time points. The proposed fault detection algorithm consists of two parts: forecasting and detection. At first we split the whole time series into equal-sized batches of length $w$ denoted as $X^{(i)} = \{x^{(j)}, x^{(j+1)}, \dots, x^{(j + w - 1)}\}$. Here $i$ is the batch number and $j = w(i-1) + 1$ is the number of first time point in the batch. In the forecasting part we predict values for the next batch $\tilde X^{(i + 1)}$ using already observed measurements $X^{(1)}, X^{(2)}, \dots, X^{(i)}$. The detection part is based on finding time points where the mean square error (MSE) between the measured $X^{(i+1)}$ and predicted $\tilde X^{(i+1)}$ values becomes higher then the precomputed threshold.

\subsection{Data Preprocessing}

All data points in the presented data set share the same time grid and have significantly varying absolute values. To reduce these variations and unify different dimensions we applied normalization transform on each dimension separately:
$$
    {x^{*}}^{(j)}_i = \frac{x^{(j)}_i - \bar x_i}{\sigma_i}, \quad i = \overline{1, m}. 
$$
Here $\bar x_i$ and $\sigma_i$ are the mean value and standard deviation for each dimension.

In the test set the additional variables labeled as ATTACK, DANGER and FAULT are introduced. They determine different parts of attack evolution. We will use the DANGER series to compare results with the fault-detection algorithm.

\subsection{Neural Network Architecture}

\begin{figure}[h]
    \centering
    \begin{minipage}{.3\textwidth}
        \centering
        \includegraphics[width=0.5\linewidth]{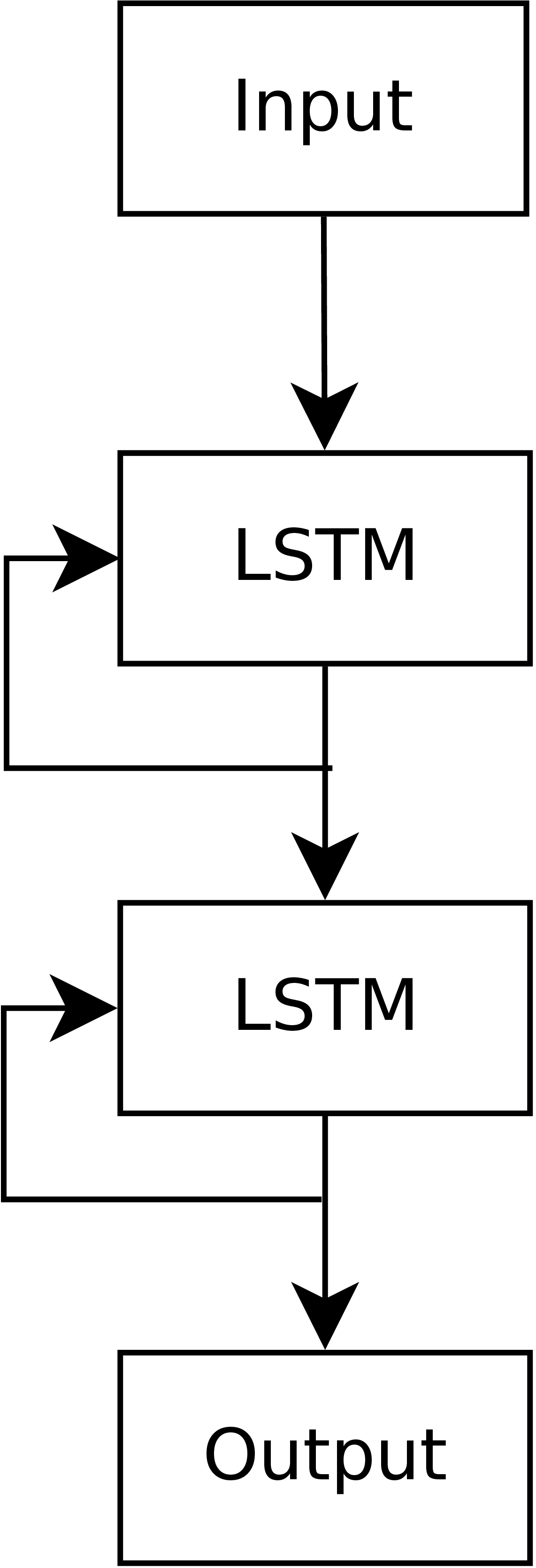}
        \caption{Neural network architecture}
        \label{fig:stacked_lstm}
    \end{minipage}
    \begin{minipage}{.6\textwidth}
        \centering
        \includegraphics[width=0.7\linewidth]{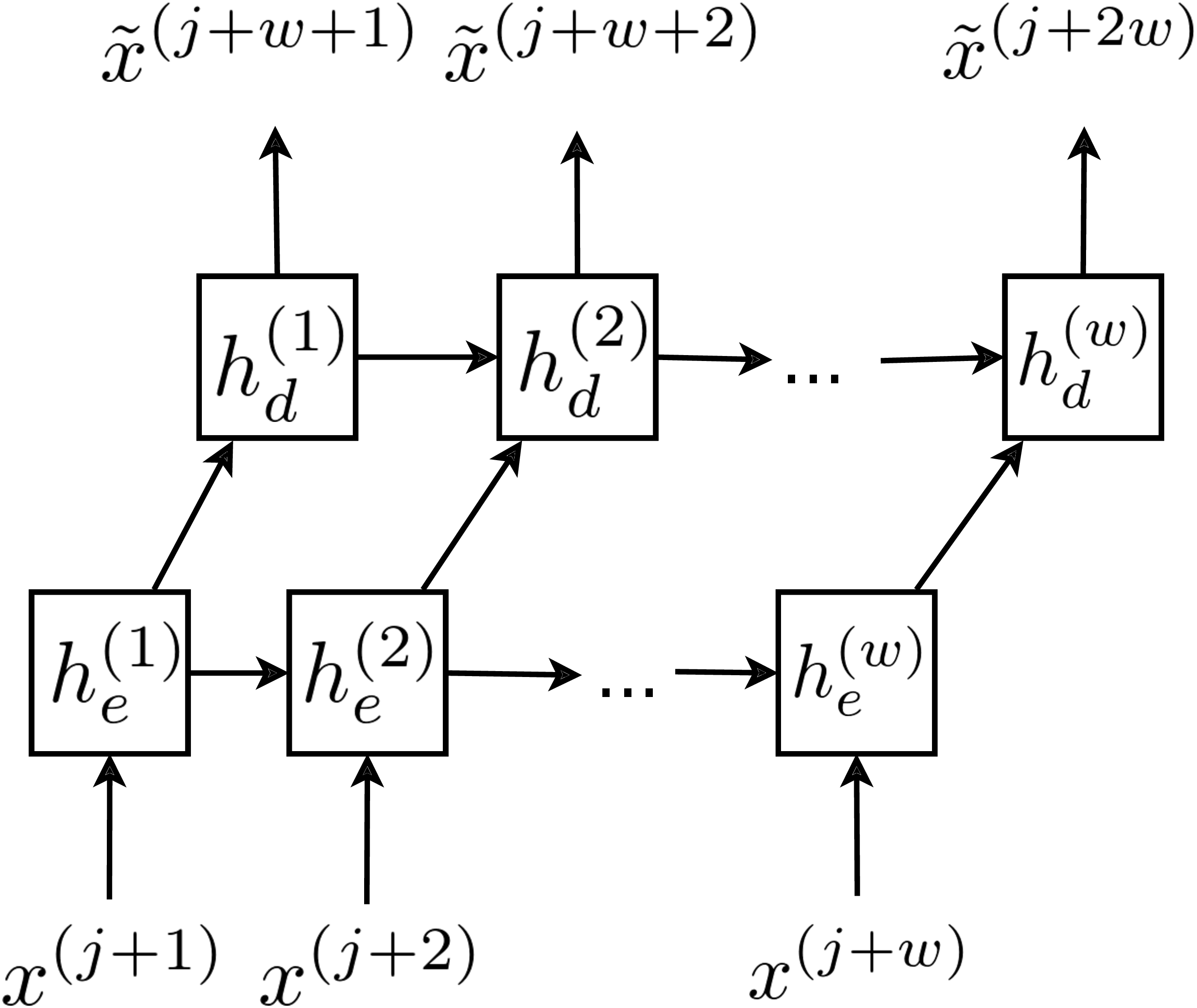}
        \caption{Forecasting scheme}
        \label{fig:lstm_ed}

    \end{minipage}
\end{figure}

The choise of optimal network architecture is based on several observations. At first, the most industrial technological processes generate strongly correlated multivariate time series. Furthermore we frequently deal with multiscale processes (see Figure~\ref{fig:data}) having fast (long-term) and slow (short-term) sub-processes. In these conditions conventional feed-forward neural networks usually demonstrate a poor results. An accurate data-driven predictive model can be developed using stateful LSTM neural network~\citep{Hochreiter:1997:LSM:1246443.1246450, Malhotra:2015, Nanduri2016AnomalyDI}. The proposed network architecture includes two stacked LSTM layers with linear output layer (Figure \ref{fig:stacked_lstm}). In addition we use a sequence-to-sequence architecture of LSTM network for the forecasting model (Figure~\ref{fig:lstm_ed}).

\begin{figure}[h]
    \centering
    \includegraphics[width=\linewidth]{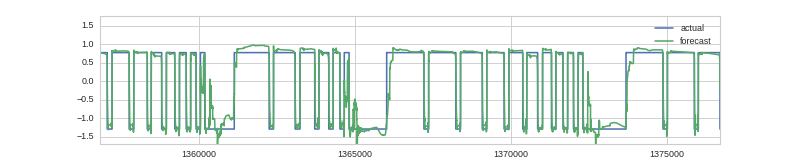}
    \caption{Example of the control variable forecast}
    \label{fig:forecast}
\end{figure}

The dropout technique~\citep{JMLR:v15:srivastava14a} is used for regularization. The results for different dropout probability values are shown in Table~\ref{table:pr}. The mean square error between training and predicted values is considered as a loss function. The RMSprop~\citep{Hinton2012} optimization algorithm is used for training. In Figure \ref{fig:forecast} an example of the forecasted values for one control variable is shown.

\begin{figure}[h]
    \centering
    \begin{minipage}{0.4\textwidth}
        \centering
        \includegraphics[width=\linewidth]{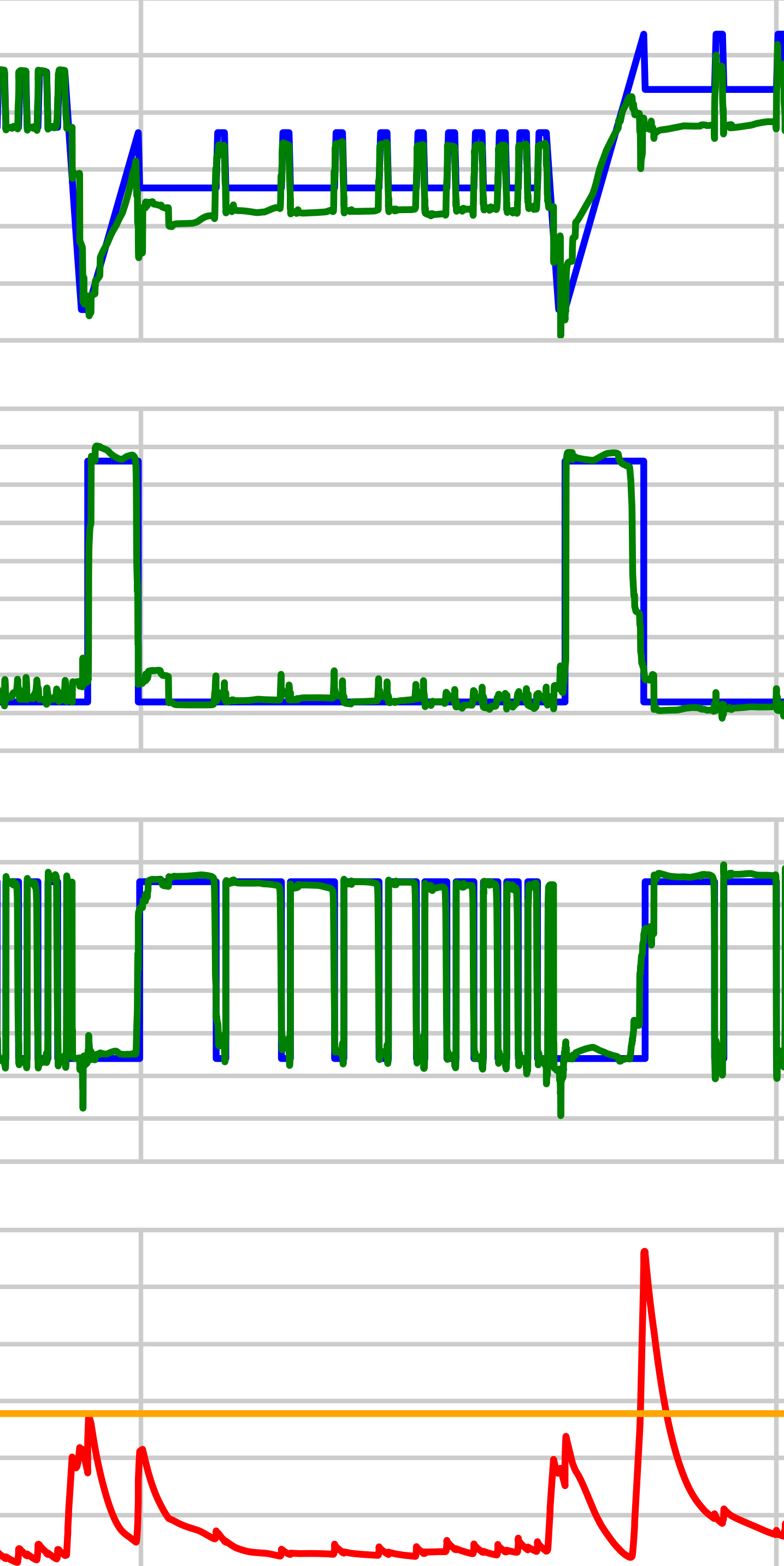}
        \caption{Example of the forecast, averaged MSE and fault detection threshold}
        \label{fig:mse}
    \end{minipage}
    \quad
    \begin{minipage}{0.5\textwidth}
        \centering
        \includegraphics[width=\linewidth]{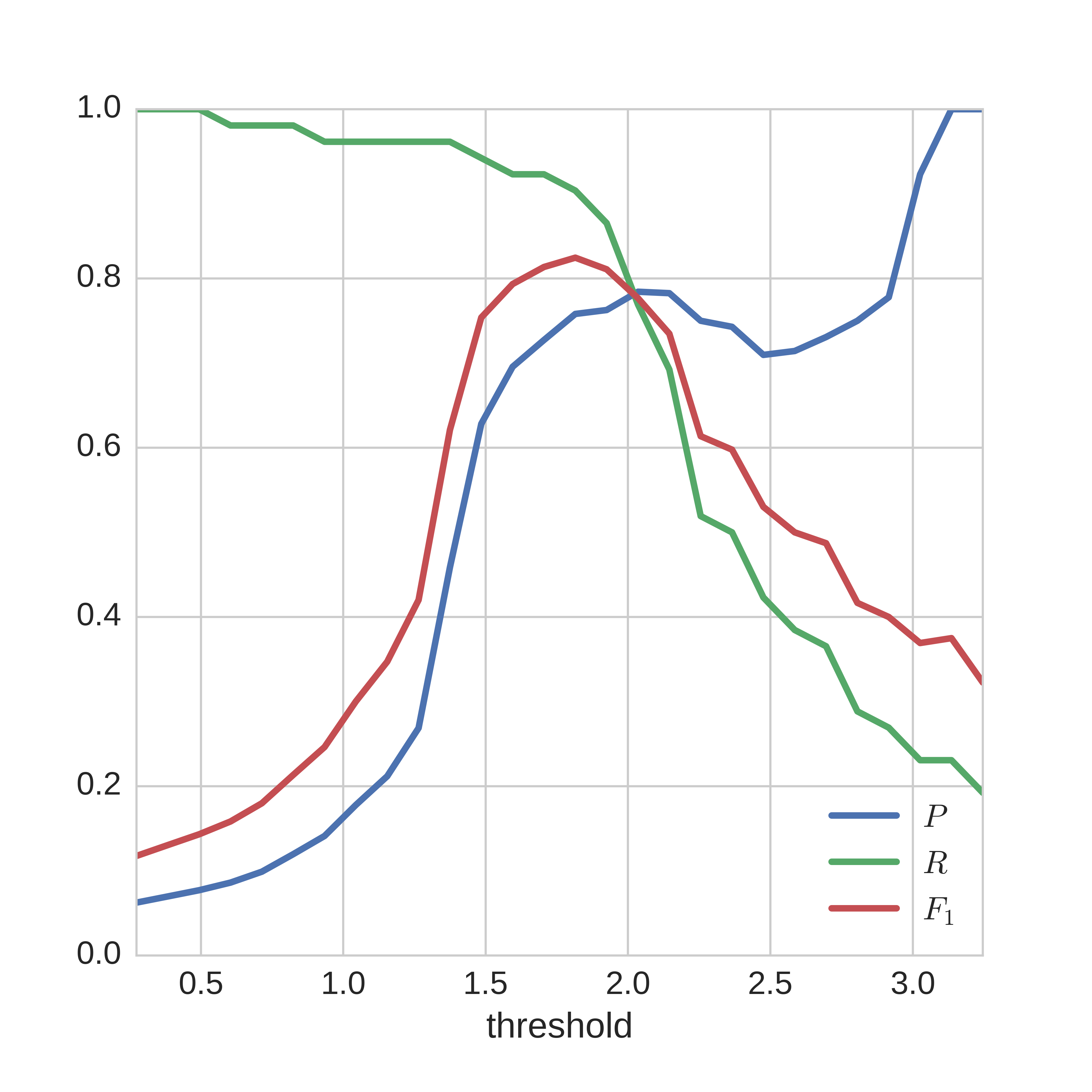}
        \caption{Precision, recall and $F_1$ score for different threshold levels}
        \label{fig:pr}
    \end{minipage}
\end{figure}

The detection part is based on the MSE between actual data and forecasted values.
$$
    MSE(X^{*}{}^{(i)}, \tilde X^{(i)}) = \frac{1}{m}\sum\limits_{i = 1}^{m}\left(x^{*}{}^{(i)}_i - \tilde x^{(i)}_i \right)^2.
$$
To smooth high errors in single points we applied an exponential moving average of MSE where the ``half-life'' exponential parameter was chosen as doubled batch length (see Figure \ref{fig:mse}). To achieve a better results in MSE computational experiments we considered only a subset of the aforementioned 19 variables. These are RT\_level, RT\_temperature, HT\_level, HT\_temperature, inj\_valve\_act and heater\_act - the most important variables partially represented in Figure~\ref{fig:data}.

According to discussion in section~\ref{sec:data_desc} we will determine process anomalies in terms of forecasting error. The horizontal line in the last subplot (Figure~\ref{fig:mse}) represents a $0.999$ quantile of empirical error distribution. This level is used as a lower boundry for the threshold in the fault-detection algorithm. The decision rule is formulated as follows: if the forecast error is less or equal to the threshold level then the algorithm indicates normal behavior and if the forecast error is greater than the threshold level the algorithm predicts abnormal behavior (fault).

\subsection{Quality Metrics}

To compute the precision and recall scores for different thresholds we split each test series into equal-sized intervals and check whether MSE is greater than the threshold level. Such a situation is treated as a fault; otherwise, an interval is classified as normal behavior.
Figure~\ref{fig:pr} illustrates how the precision, recall and $F_1$ scores depend on threshold level. An interesting practical aspect of the results represented in Figure~\ref{fig:pr} is that the threshold level may be used as a tunable parameter that can be changed to achieve desired fault positive rate. This aspect can help us to handle the problem of lots of false positive alerts in a monitoring system. The operator of an industrial object can set this parameter to suitable level.

The best $F_1$ score results for different batch size~($w$) and dropout probability~($p$) are represented in Table~\ref{table:pr}
\begin{table}[h]
    \centering
    \begin{minipage}{0.45\linewidth}
    \centering
    \begin{tabular}{|c|c|c|c|c|c|}
        \hline
         $\mathbf w$ & $\mathbf MSE$ & \bf Precision & \bf Recall & $\mathbf F_1$ \\
        \hline
        \hline
        $30$ & $0.318$ & $0.450$ & $0.346$ & $0.391$ \\
        \hline
        $60$ & $0.124$ & $0.632$ & $0.462$ & $0.533$ \\
        \hline
        $90$ & $0.227$ & $0.732$ & $0.788$ & $0.759$ \\
        \hline
        $120$ & $0.194$ & $0.782$ & $0.827$ & $\bf 0.804$ \\
        \hline
        $150$ & $0.230$ & $0.683$ & $0.788$ & $0.732$ \\
        \hline
        $180$ & $0.203$ & $0.585$ & $0.923$ & $0.716$ \\
        \hline
    \end{tabular}
    \end{minipage}
    \begin{minipage}{.45\linewidth}
    \centering
    \begin{tabular}{|c|c|c|c|c|}
        \hline
        $\mathbf p$ & \bf Precision & \bf Recall & $\mathbf F_1$ \\
        \hline
        \hline
        $0.5$ & $0.782$ & $0.827$ & $0.804$ \\
        \hline
        $0.1$ & $0.976$ & $0.788$ & $\bf 0.872$ \\
        $0.01$ & $0.846$ & $0.846$ & $0.846$ \\
        \hline

    \end{tabular}
    \end{minipage}
    \caption{Results of experiments}
    \label{table:pr}
\end{table}

\subsection{Comparison With Other Methods}

The most known methods of industrial fault detection are given in~\citep{0957-0233-12-10-706}. Table~\ref{table:comparsion} shows comparision results of conventional fault detection methods versus the proposed approach tested at 6 aforementioned variables. 

\begin{table}[h]
    \centering
    \begin{tabular}{|c|c|c|c|}
    \hline
     \bf Method & \bf Precision & \bf Recall & $\mathbf F_1$ \\
    \hline
    \hline
     LSTM & $0.976$ & $0.788$ & $\bf 0.872$  \\
     PCA & $0.750$ & $0.611$ & $0.673$ \\
     FDA & $0.909$ & $0.185$ & $0.308$ \\
     PLS & $1.000$ & $0.426$ & $0.597$ \\
     CVA & $0.968$ & $0.556$ & $0.706$ \\
     OCSVM & $0.422$ & $0.885$ & $0.571$ \\
    \hline
    \end{tabular}
    \caption{Results of methods comparision}
    \label{table:comparsion}
\end{table}

As it follows from table~\ref{table:comparsion} such methods as PCA, FDA and СVA show good results in precision but not in recall. The OneClassSVM with radial basis functions as the kernel achieves the best recall but poor precision. The PCA and LSTM show balanced results in both metrics. The LSTM dominates PCA and achieves best averaged ($F_1$) result for described dataset.

\section{Conclusion and Future Work}
\label{seq:conc}

The current paper presents a publicly available dataset for the problem of industrial fault detection. This dataset consists of a multivariate time-series training set and dozens of test sets with different types of faults. Like the Tennessee Eastman process~\citep{TEChallenge}, the  proposed dataset includes both sensor and control, continuous and  discreets channels for analysis. The results obtained in section~\ref{seq:lstm_fd} show that the LSTM-based fault-detection approach has advantages over classic fault-detection methods~\citep{0957-0233-12-10-706}. The error threshold level was introduced as a tunable parameter that allows a user to achieve a satisfactory false positive and false negative detection rate.

The fault-detection approach described in section~\ref{seq:lstm_fd} restricts us to a binary decision: the system either operates in normal or abnormal mode. From a practical point of view, such a system has the  following disadvantages: alerts cannot be prioritized and interpreted. A possible modification to the proposed approach is to add strict order. Some kind of abnormality measure may help to prioritize alerts triggered by a monitoring system. Such measure may also provide the possibility to use a more complex quality metric such as receiver operating characteristic (ROC). Another possible modification is to add methods for fault diagnosis~\citep{0957-0233-12-10-706} to provide not only the moment of time when a fault is detected but also to localize the subset of channels where it was detected. This problem is particularly important in the analysis of large dimension time series.

Another research direction we see in GHL-model improvement in order to reach more realistic data via   including stochastic parameters,  measurement noise and random outlayers. This will enrich process trajectories and allows us to test low-order statistical parametric model and change point techniques.

\end{document}